\documentclass[conference]{IEEEtran}
\IEEEoverridecommandlockouts
\usepackage{cite}
\usepackage{amsmath,amssymb,amsfonts}
\usepackage{algorithmic}
\usepackage{graphicx}
\usepackage{textcomp}
\usepackage{xcolor}

\usepackage[switch]{lineno}

\usepackage{siunitx}
\usepackage[mode=buildnew]{standalone}

\definecolor{reviewblue}{RGB}{0,50,180}
\newcommand{\rev}{} 

\begin{document}

\title{Adding numbers with spiking neural circuits on neuromorphic hardware: A building block for future hybrid systems\\
}

\author{\IEEEauthorblockN{Oskar von Seeler}
\IEEEauthorblockA{\textit{Dept. of Neuro- and Sensory Physiology} \\
\textit{University Medical Center Göttingen}\\
Göttingen, Germany \\
oskar.vonseeler@stud.uni-goettingen.de}
\and
\IEEEauthorblockN{Elena C. Offenberg}
\IEEEauthorblockA{\textit{Third Institute of Physics -- Biophysics} \\
\textit{University of Göttingen}\\
Göttingen, Germany \\
e.c.offenberg@umcutrecht.nl}
\and
\IEEEauthorblockN{Carlo Michaelis}
\IEEEauthorblockA{\textit{Third Institute of Physics -- Biophysics} \\
\textit{University of Göttingen}\\
Göttingen, Germany \\
carlo@evocracy.org}
\and
\IEEEauthorblockN{Jannik Luboeinski\textsuperscript{\#}\textsuperscript{*}}
\IEEEauthorblockA{\textit{Dept. of Neuro- and Sensory Physiology} \\
\textit{University Medical Center Göttingen}\\
Göttingen, Germany \\
jannik.luboeinski@med.uni-goettingen.de}
\and
\IEEEauthorblockN{Andrew B. Lehr\textsuperscript{\#}}
\IEEEauthorblockA{\textit{Dept. of Neuro- and Sensory Physiology} \\
\textit{University Medical Center Göttingen}\\
Göttingen, Germany \\
andrew.lehr@med.uni-goettingen.de}
\and
\IEEEauthorblockN{Christian Tetzlaff\textsuperscript{\#}}
\IEEEauthorblockA{\textit{Dept. of Neuro- and Sensory Physiology} \\
\textit{University Medical Center Göttingen}\\
Göttingen, Germany \\
christian.tetzlaff@med.uni-goettingen.de}
}

\maketitle

\null
\vspace{-1.2cm}
\noindent \textsuperscript{\#}co-senior authors \quad \textsuperscript{*}corresponding author
\vspace{0.3cm}

\begin{abstract}
Progress in neuromorphic computing requires efficient implementation of standard computational problems, like adding numbers.
Here we implement \rev{a variety of} sequential and parallel binary adders in the Lava software framework, and deploy them to the neuromorphic chip Loihi 2. \rev{To the best of our knowledge, up to now, a neuromorphic implementation of such parallel adders has not been reported.}
We describe the time complexity, neuron and synaptic resources, as well as constraints on the bit width of the numbers that can be added with the current implementations. Further, we measure the time required for the addition operation on-chip.
Importantly, we encounter trade-offs in terms of time complexity and required chip resources for the three considered adders.
While sequential adders have linear time complexity \boldmath$\mathcal{O}(n)$ and require a linearly increasing number of neurons and synapses with number of bits $n$, the parallel adders have constant time complexity \boldmath$\mathcal{O}(1)$ and also require a linearly increasing number of neurons, but nonlinearly increasing synaptic resources (scaling with \boldmath$n^2$ or \boldmath$n \sqrt{n}$).
This trade-off between compute time and chip resources may inform decisions in application development, and the implementations we provide may serve as a building block for further progress towards efficient neuromorphic algorithms.
\end{abstract}


\section{Introduction}\label{sec.introduction}

The promise of neuromorphic hardware lies in highly parallel and energy-efficient representations and computation, taking inspiration from networks of neurons in the brain \cite{davies2021advancing,schuman2022opportunities}.  
However, not all tasks lend themselves to decentralized, parallelized implementations, meaning that standard computation will be required in conjunction with -- and not replaced by -- neuro-inspired neuromorphic algorithms.
Future computing systems will thus often be of hybrid nature, depending on the integration of CPUs and GPUs, related memory systems, as well as novel decentralized chip architectures like neuromorphic hardware.

As for any computing architecture, transferring data between systems is a bottleneck, which also holds true for neuromorphic chips. 
This raises a critical question: Should a system transfer data from a neuromorphic chip to standard hardware to perform certain computations and then transfer this processed data back to the neuromorphic chip?
Or would it overall be more efficient to develop decentralized implementations of standard functionality to be used on the neuromorphic system, in order to avoid the data transfer bottleneck (despite the implementation itself potentially being slower and/or less efficient)?
\\
\indent The answer to this question ultimately depends on the details of the considered system.
However, when the amount of data to transfer becomes large enough, an on-chip implementation of standard computational operations is likely to outperform solutions requiring data transfer.
For example, in a large-scale neuro-inspired architecture, it may be more prudent to implement a cost function for optimization, or a simple procedure for data analysis, using arithmetic operations directly on chip instead of transferring data back and forth to and from a CPU.
\\
\indent Given this issue, we set out towards on-chip implementations of standard operations with the goal of minimizing unnecessary data transfer.
To start in this direction, here we consider addition, one of the fundamental arithmetic operations required by all modern computing applications. Addition is needed for example when computing error functions for online learning~\cite{sussillo_generating_2009,ponulak2010supervised,zenke2018superspike,bellec2020solution}, or when calculating shortest paths in graph search algorithms that are optimally suited for parallelized implementation on neuromorphic hardware~\cite{aimone_provable_2021}.

Here, we contribute to the set of computational elements for neuromorphic hardware \emph{a set of binary adders}, implemented using circuits of spiking neurons in Lava.
This enables us to test the adders on the neuromorphic chip Loihi 2 and to provide a basis for future implementations on other neuromorphic systems.
As a major outcome of our investigations, we find that there are trade-offs between the different adders with respect to their time complexity and required number of neurons and synapses. In this context, we characterize the bit precision enabled by the different algorithms and demonstrate under which circumstances an algorithm makes efficient use of compute resources.

\section{Background}\label{sec.background}

\subsection*{Classical algorithms for integer addition}
For digital computers, adder circuits naturally operate on binary numbers. Typical building blocks of adder circuits are half and full adders \cite{HennessyPatterson2012}. A half adder takes two bits as inputs and produces a sum bit and a carry bit. A full adder extends this by including a carry input.

A simple multiple bit adder design is the ripple-carry adder \cite{HennessyPatterson2012}, which connects multiple full adders in sequence, passing the generated carry from one full adder to the carry input of the next full adder. Since the time taken by the ripple-carry adder grows linearly with the number of bits \cite{HennessyPatterson2012}, other designs are often used in practice.

One alternative is the carry-lookahead adder \cite{HennessyPatterson2012}, which uses half adders at the inputs for all bit positions. The resulting sum and carry out values indicate whether an incoming carry is propagated or if a carry will be generated at that position. Using these signals, the computation can be split into groups that perform parallel computation with fast processing of carry signals. 

\subsection*{Neuromorphic algorithms for integer addition}

Various designs for addition circuits can be found in threshold logic research. A threshold gate, the building block of threshold logic networks, takes a weighted sum of inputs and produces a signal if that sum exceeds a set threshold. These gates are more flexible than traditional logic gates used in electronic circuits, allowing for different designs of adder circuits~\cite{ramos_two_1999,siu_depth-size_1991}.
Ramos and Boh\'orquez introduced a family of ``double carry threshold adders'' (DCTA), defining adders for any fixed depth $d \ge 2$~\cite{ramos_two_1999}. In neuromorphic implementations, a fixed depth can be implemented by a fixed number of time steps for computation, independent of the number of bits.
From the family of adders presented in \cite{ramos_two_1999}, here, we focus on the DCTA2 and DCTA3 algorithms.

Motivated by the progress in the field of neuromorphic computing, recent research has explored integer adders specifically designed for current neuromorphic hardware systems.
Binary encoding of integers for addition has been shown to be efficient and balanced regarding time, neurons, synapses and energy consumption when compared to binning, rate-based or time-based encodings~\cite{date_encoding_2023}.
Recently, multiple binary adder designs have been created and tested on neuromorphic hardware emulators \cite{date_encoding_2023,wurm_arithmetic_2023} and on hardware~\cite{aimone_spiking_2022,iaroshenko2023binary,ayuso-martinez_analog_2025}. These designs enable efficient use of neurons and synapses under specific circumstances. Furthermore, Date et al.~\cite{date_encoding_2023} extended simple binary encoding to include additional bits for negative and rational parts, and Wurm et al.~\cite{wurm_arithmetic_2023} used the established \textit{two's complement} method to encode signed integers. However, each of these solutions operate in a sequential manner, meaning that the number of required time steps depends on the number of bits that are considered.
To the best of our knowledge, up to now, a neuromorphic implementation of parallel adders as proposed by \cite{ramos_two_1999} has not been reported.

Aimone et al. implemented a streaming adder that processes one bit at a time using the same neurons and synapses, and recurrent connections for the carry signal~\cite{aimone_spiking_2022}. This leads to a resource efficient implementation when the streaming format is applicable. If the sum has to be represented using multiple neurons at the same time step, converting the stream to that format requires further resources. 

\subsection*{Benefits and constraints of the Loihi 2 neuromorphic chip}
%

The Loihi 2 chip by Intel includes 128 neuromorphic cores with 1 million neuron units in total, connected by a network-on-chip of up to 120 million synapses. It further contains 6 embedded microprocessor cores that enable network configuration and monitoring. Using this architecture, Loihi 2 enables computation with spiking neurons that communicate asynchronously via on-chip synapses, and it exhibits efficient processing of temporal dynamics in sparsely connected networks \cite{davies2021advancing,timcheck2023intel}.

Loihi 2 further supports the implementation of custom neuron models by a set of specific microcode commands (similar to assembler code) and allows for delayed and graded spikes. It also provides a variety of mechanisms to implement synaptic plasticity rules. As we are neither using a custom neuron model nor synaptic plasticity in the present work, we shall not discuss these features in more detail, however, they may be beneficial for future applications or enhancements of this work. In particular, microcoded neurons might provide means to further boost the performance of the discussed adders, but at the loss of generality for other neuromorphic systems. Finally, a very powerful feature of Loihi 2 is its scalability in 2D and 3D. Currently, systems with more than 1000 Loihi 2 chips are already available \cite{hala-point-news}.

While the features of the Loihi 2 chip are intended to be representative for future commercial neuromorphic devices \cite{timcheck2023intel}, it also comes with natural constraints, posing issues that need to be addressed in the development of neuromorphic algorithms.
First of all, there is overhead required for initializing and monitoring networks on chip that must be taken into account during algorithm design.
Furthermore, model variables and parameters like synaptic weights and delays can only take values with a certain, fixed precision.
In this study, we present neuromorphic solutions that cope with these constraints and demonstrate how different adders can be implemented efficiently on the Loihi 2 chip.

Together with Loihi 2, Intel launched the open-source framework Lava \cite{lava-website}, which aims to provide a common framework for neuro-inspired applications on different backend architectures. The major aspects of Lava are asynchronous communication via spikes, exchangeable backends such as CPU and Loihi 2 models, modularity, and a Python API.

\section{Methods}\label{sec.methods}

In our algorithmic implementations, the integers to be added are represented via bits, and these bits are represented in Lava and on Loihi 2 by spikes or the absence of spikes at specified time steps. We assume that the operands as well as the result fit within a fixed bit width $n$, which is specified when the adder is created.
Overflows can be detected by the spiking of the carry of the most significant bit.
The adders that we consider take all of their input bits simultaneously, and produce all output spikes at once with no additional spikes at different time steps. We are particularly interested in highly parallel constant time adders, and implement DCTA2, DCTA3 and one linear time adder as a comparison.

The different adder circuits are implemented using the Lava framework \cite{lava-website}. In the terminology of this framework, an adder is a \texttt{SubProcess} with two incoming ports for the two operands and one outgoing port for the result. The incoming ports receive spikes as input and the outgoing port produces spikes as outputs at a specific time after receiving the input spikes. The setups of the adders only depend on the bit width they should support.

The adders are implemented using the leaky integrate-and-fire (LIF) neuron model.
Groups of neurons are connected using a \texttt{Sparse} connections object to avoid creating superfluous zero-weight synapses on the chip. 
Decay values are set such that membrane voltages and currents decay to 0 by the next time step to avoid any residuals affecting the next computations. Bias and firing threshold are adjusted individually.

With these parameters, the neurons act as simple threshold gates that fire when the sum of the weighted synaptic input currents exceeds the firing threshold within a single time step.
We denote $n$-bit numbers as sequences of bits $X \in \{0,1\}^n$, with $X_0$ being the least significant and $X_{n-1}$ the most significant bit.
In the following, we describe the process of adding two $n$-bit numbers $X$ and $Y$, yielding the sum $S \in \{0,1\}^n$, for each type of adder.
%

All adders compute a sum bit $S_i$ using the two input bits $X_i$ and $Y_i$, the carry in $C_{i-1}$, and the carry out $C_i$:

\newcommand{\Threshold}[1]{\begin{cases}
		1, & \text{if $#1$,}\\
        0, & \text{otherwise.}
\end{cases}}

\begin{equation}
    S_i = \Threshold{X_i+Y_i+C_{i-1}-2C_i \ge 1}
\end{equation}

The sum bit is $1$ if (i) any of the three input bits is $1$ and the carry out is $0$, or (ii) if all three inputs and the carry out are $1$.
This is implemented using a LIF neuron with threshold $1$ 
and weighted synaptic connections with weight $1$ for the input bits $X_i$, $Y_i$ and the carry in bit $C_{i-1}$ and weight $-2$ for the carry out bit.

We have implemented three adder circuits, with different methods to compute the carry. The first one is a sequential adder inspired by ripple-carry adders,
the other two are the fully parallel adders DCTA2 and DCTA3~\cite{ramos_two_1999}.

\subsection{Sequential adder}

\begin{figure}[htbp]
    \centering
    \includegraphics{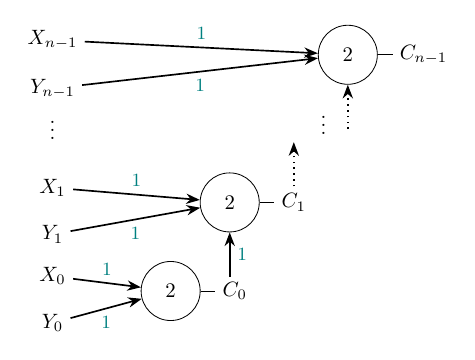}
    \caption{Carry computation of the sequential adder. Neurons are shown as circles with their firing threshold. Synapses are drawn as arrows with their weights in green. Synaptic delays are indicated by length of arrows.}
    \label{fig:seq-adder}
\end{figure}

The sequential adder computes one bit at a time (cf. Fig.~\ref{fig:seq-adder}). Having the carry in available allows for a straightforward calculation of the carry out:

\begin{equation}
    C_i = \Threshold{C_{i-1}+X_i+Y_i \ge 2}
\end{equation}

As the carry is computed with a single neuron and is propagated to the next carry neuron immediately, the adder requires one additional time step per bit, and one further step to compute all the sums.
To allow input and output bits to be synchronized, the implementation of the sequential adder makes use of synaptic delays.
As Loihi 2 has a maximum synaptic delay of 63 time steps \cite{orchard2022talk}, and the input spikes reach the summing neurons with a delay of exactly $n$ steps, this adder only works up to $63$ bits of precision.
If higher precision is required, a relay neuron with threshold $0$ can be inserted in place of each synapse. This adds one time step of delay for the processing in the neuron and up to another $63$ time steps of delay through the second synapse, therefore increasing the delay and maximum precision by $64$ bits for each added layer of relay neurons.

\subsection{DCTA2 parallel adder}

\begin{figure}[htbp]
    \centering
    \includegraphics{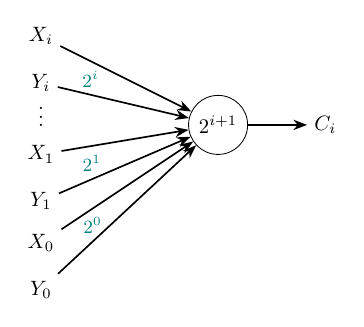}
    \caption{Computation of the carry for a single bit position in the DCTA2 adder. The circle is a single neuron with its firing threshold. Synapses are drawn as arrows to and from the neuron, with weights indicated between synapses.}
    \label{fig:dcta2-adder}
\end{figure}

Avoiding the linear time needed by the sequential adder, the DCTA2 adder computes all carry bits in parallel~\cite{ramos_two_1999}. Constrained to depth $2$,
each carry value has to be computed from the input bits within a single neuron:

\begin{equation}
    C_i = \Threshold{\sum_{j=0}^{i} 2^{j} X_{j}+2^{j} Y_{j} \ge 2^{i+1}}
\end{equation}

Compared to the sequential adder that only uses weights of $1$ and $-2$, DCTA2s synaptic weights grow exponentially with the number of bits to be added.
Our implementation supports DCTA2 with up to 16 bits of precision, limited by these exponential weights.
A single synapse group on Loihi allows for 8 bits of precision in the weight mantissas. To achieve 16 bits, we add a second synapse group with the weight exponent increased by 8, allowing for another 8 bits of weight precision. 
A third synapse group cannot be used for a further increase of precision because this would exceed the limit of the weight exponent allowed by Loihi 2.

\subsection{DCTA3 parallel adder}

\begin{figure}[htbp]
    \centering
    \includegraphics[width=0.48\textwidth]{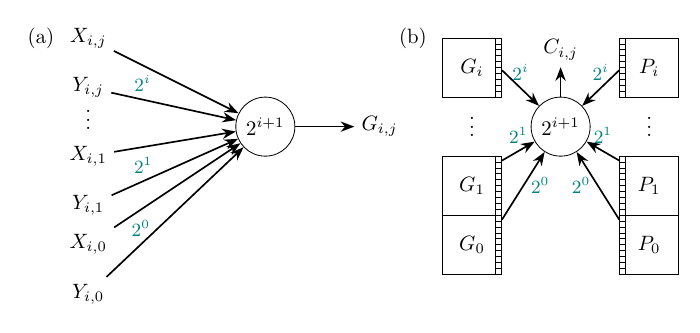}
    \caption{Computation of the \emph{generate}/\emph{propagate} and \emph{carry} signals for a single bit position in the DCTA3 adder. \textbf{(a)} The sketch on the left shows the generate signal $G_{i,j}$ -- the computation is the same for a \emph{propagate} signal, but with an adjusted threshold. \textbf{(b)} Groups of generate and propagate signals are shown as large boxes, with the individual signals indicated as small squares. Groups and signals within each group are ordered with the most significant positions at the top. The carry neuron takes the generate and propagate signals of the most significant bit of each less significant group, and the generate and propagate signal from the same position, to determine the carry.}
    \label{fig:dcta3-adder}
\end{figure}

Increasing the depth to $3$ allows computing the carry with slightly smaller weights to enable adding larger integers~\cite{ramos_two_1999}.
This adder design is similar to carry-lookahead adders.
The computation is split into $\left\lceil \sqrt n \right\rceil$ groups of up to $\left\lceil n / \sqrt n \right\rceil$ bits each.
Each group is then treated as a separate addition and carries are computed as in DCTA2. These computed carries are not necessarily the same as the carries for the full addition, they are intermediate values referred to as the \emph{generate} signal:

\begin{equation}
    G_{i,j} = \Threshold{\sum_{k=0}^j 2^k X_{i,k} + 2^k Y_{i,k} \ge 2^{j+1}}
\end{equation}
Here, $i$ is the group and $j$ is the index of the bit within the group. Additionally, the \emph{propagate} signal is needed to compute the carry. For a specific position within a group, this signal indicates whether a carry would occur if the group receives an incoming carry at its least significant position. Instead of passing an additional spike with weight 1 to the \emph{propagate} threshold gate, the threshold can be reduced by 1 compared to the generate computation:
\begin{equation}
    P_{i,j} = \Threshold{\sum_{m=0}^j 2^m X_{i,m} + 2^m Y_{i,m} \ge 2^{j+1}-1}
\end{equation}

From generate and propagate values at a specific index, together with generate and propagate values from prior groups, the actual carry can be computed.

\begin{equation}
    C_{i,j} = \Threshold{GP_{i,j} \ge 2^{i+1}}
\end{equation}
where $GP_{i,j} = 2^i G_{i,j} + 2^i P_{i,j} + \sum_{k=0}^{i-1} 2^k G_{k,-1} + 2^k P_{k,-1}$ ($G_{k,-1}$ refers to the generate signal at the most significant position within group $k$).
The individual threshold gates for propagate, generate and carry all have different threshold values depending on their position. However, Lava does not allow for different thresholds within a single neuron group. To avoid creating many neuron groups (that would each run on their own core), we use bias input to offset the threshold. With the current implementation, the available bias precision on Loihi 2 limits the DCTA3 to adding up to $42$ bits. However, this could be extended by setting individual thresholds for each neuron. Without shared thresholds, the adder would be limited by synaptic weight precision in the same way as DCTA2. With 16 bits of precision, the adder would work for up to 16 groups of 16 bits each, for a total of 256 bits.

\section{Results}\label{sec.results}

\begin{figure}[htbp]
\centerline{\includegraphics[width=\linewidth]{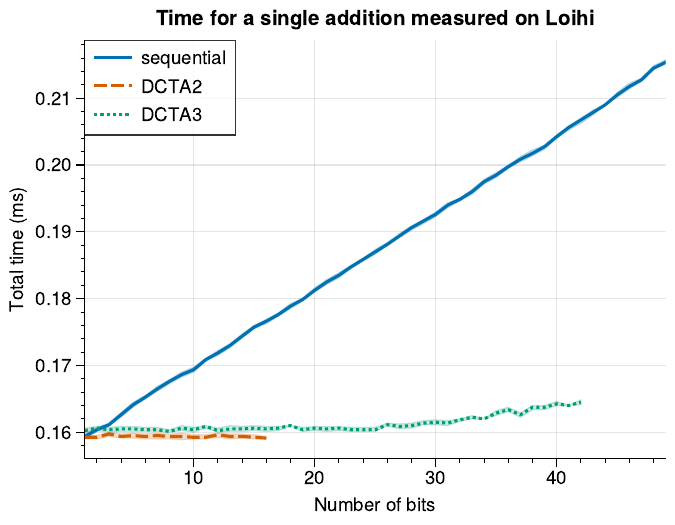}}
\caption{Runtime measured on Loihi 2 for a single addition. Compared to the theoretical time steps given in \textit{Table~\ref{tab1}}, this includes two additional steps for producing input spikes and collecting outputs. Measured times are averaged over 8 runs for each combination of adder and number of bits. The DCTA2 and DCTA3 adders are run within their supported precision range given in \textit{Table~\ref{tab1}}. Shaded area shows 95\% confidence interval.}
\label{fig:totaltime}
\end{figure}

\begin{figure}[htbp]
\centerline{\includegraphics[width=\linewidth]{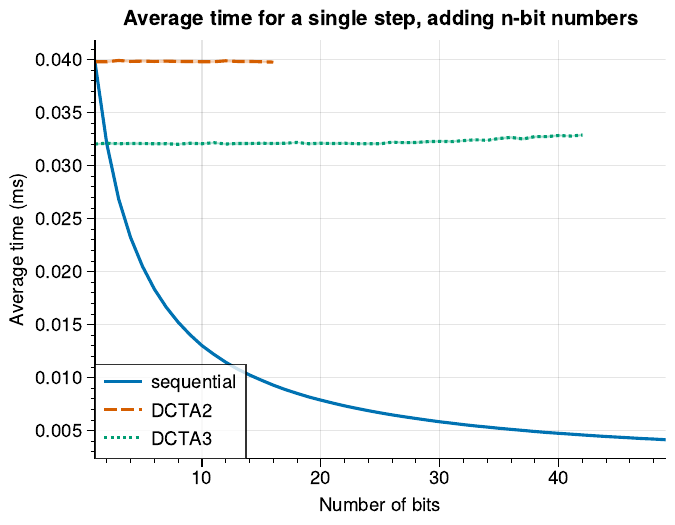}}
\caption{Average time taken for a single time step during a single addition. Runtime measured on Loihi 2, for details on the testing protocol see Fig.~\ref{fig:totaltime}.}
\label{fig:avgtime}
\end{figure}

We implemented each of the described binary adders as a \texttt{SubProcess} in the Lava framework and deployed them to an OheoGulch Loihi 2 system (with chip version N3B3 and Lava-Loihi v0.7.0).
We first provide an overview of the time complexity, the required neurons and synapses, as well as the restrictions on the bit precision of numbers that can be processed by each adder (\textit{Table~\ref{tab1}}).
Then, by adding numbers on Loihi 2 with each of the implemented binary adders, we demonstrate how the runtime scales with the number of bits in each to-be-added number
(Figs. \ref{fig:totaltime} and \ref{fig:avgtime}).

\begin{table}[htbp]
\caption{Various theoretical metrics of different adders}
\begin{center}
\begin{tabular}{|c|c|c|c|c|}
    \hline
    \textbf{Adder} & \textbf{Time} & \textbf{Neurons} & \textbf{Synapses} & \textbf{max. bits} \\ \hline
    \rev{VN}\cite{date_encoding_2023,wurm_arithmetic_2023}$^{\mathrm{a}}$ & $n+1$ & $4n-1$ & $12n-6$ & $63$ \\ \hline
    streaming \cite{aimone_spiking_2022} & $n+1$ & $4$ & $9$ & $\infty$ \\ \hline
    sequential & $n+1$ & $2n$ & $7n-2$ & $62$ \\ \hline
    DCTA2 & $2$ & $2n$ & $n^2 + 5n - 1$ & $16$ \\ \hline
    DCTA3 & $3$ & $4n$ & $3n\sqrt n+7n-1^{\mathrm{b}}$ & $42$ \\ \hline
    \multicolumn{5}{l}{$^{\mathrm{a}}$Adjusted for our task.} \\
    \multicolumn{5}{l}{$^{\mathrm{b}}$If number of bits $n$ is a perfect square. Close estimate otherwise.}
\end{tabular}
\label{tab1}
\end{center}
\end{table}

\begin{figure}[htbp]
\centerline{\includegraphics[width=\linewidth]{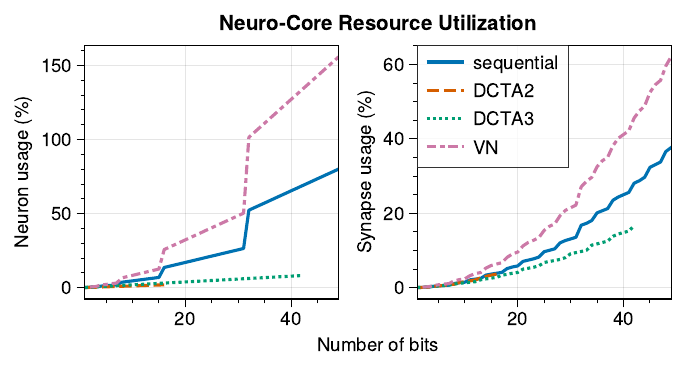}}
\caption{\rev{Same experiments as in Fig.~\ref{fig:totaltime}. Chip resource usage as measured by Lava, as a fraction of a single neuromorphic core. This includes neurons for providing input and neurons for reading output. The jumps in neuron usage occur at powers of 2, where an additional bit for the maximum synaptic delays halves the neuron capacity on Loihi 2.}}
\label{fig:resusage}
\end{figure}

The \rev{`virtual neuron' (VN) adder from }references~\cite{date_encoding_2023,wurm_arithmetic_2023}, which we consider for comparison, has solved a slightly different problems. They allow for adding integers of different bit length, have a separate adder for adding positive and negative parts, or include an additional bit in the output to account for potential overflow. For the comparison in \textit{Table~\ref{tab1}} we removed all neurons and synapses that were not necessary for adding two numbers of $n$ bits with an $n$ bit result. \rev{The streaming adder~\cite{aimone_spiking_2022} uses a different format for input and output than the other four adders. Inputs are read, starting with the least significant bit, one bit in each timestep through the same synapse and output is produced in the same way. This allows for reusing neurons for the computation across bits, leading to a circuit size that is independent of the number of bits in the input.} While the table gives theoretical neuron and synapse counts, actual resource usage may be different, depending on the hardware platform. For example, on Loihi 2, the number of available neurons is reduced if longer delays are used \cite{orchard2022talk}. \rev{We measured the actual resource usage for our three target adders, as well as for an implementation of the VN adder, on Loihi 2 and found that every one of our target adders, within its supported range, fits on a single neuromorphic core (Fig.~\ref{fig:resusage}).} This comparison reveals that DCTA3 uses less resources on Loihi 2 than the sequential adder because of lower delays, even though the theoretical metrics in Table~\ref{tab1} suggest that DCTA3 uses more neurons and synapses.
\rev{When comparing the sequential adder to the other linear time adder VN, we can see that the theoretical reduction in synapses and neurons translates directly to a reduction in resource usage on the Loihi 2 hardware (Fig.~\ref{fig:resusage}), as both designs utilize similar synaptic delays to synchronize inputs and outputs.}

\begin{figure}[htbp]
\centerline{\includegraphics[width=\linewidth]{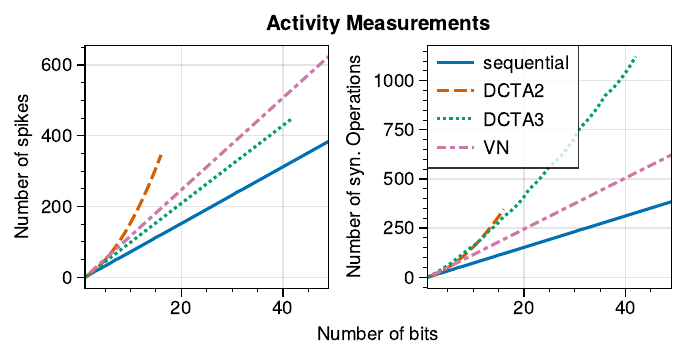}}
\caption{\rev{Spikes and synaptic events measured using a Lava \textit{Profiler}. Input and output spikes are included. For the sequential adder and DCTA2, the two measures are identical, but for DCTA3 there are more synaptic events than spikes.}}
\label{fig:actcount}
\end{figure}

To experimentally evaluate runtime results of the adders, a calculation maximizing the number of spikes is selected. For a given number of bits $n$, we add $2^{n-1}-1$ to itself, resulting in $2^n-2$. To provide input spikes and read the outputs, three additional groups of $n$ neurons are connected to the ports of the adder component. In addition to the time steps provided in \textit{Table~\ref{tab1}}, the execution runs for two more steps to provide the input and extract the output. The execution time is measured on chip using the \texttt{Loihi2HWProfiler} class provided by Lava. Results are shown in Fig.~\ref{fig:totaltime}. Within the tested number of bits the runtime for the sequential adder scales linearly with a slope of \SI{\sim 1.1}{\micro\second/bit}.
The implementation of the DCTA2 adder can handle numbers between $1$ and $16$ bits (see \textit{Table \ref{tab1}}), and within that region, the runtime remains constant.
DCTA3 takes one more time step compared to DCTA2 and is therefore marginally slower. When scaling beyond $30$ bits, the runtime begins to increase slightly. This might be due to the high number of synaptic operations this adder requires (see Fig.~\ref{fig:actcount}).
\rev{While the number of synaptic operations might grow more quickly for DCTA2, once the runtime of DCTA3 starts to increase at about 30 bits (see Fig. \ref{fig:totaltime}), the number of operations is higher than that of the largest sequential and DCTA2 adders we have tested.
In terms of spikes and synaptic operations, the sequential adder is most efficient among the adders we have implemented.}

From our runtime measurements with the Loihi 2 profiler, we find that there is an overhead that does not depend on the total number of time steps. The overhead is visible in the shape of the sequential adder's average step time in Fig.~\ref{fig:avgtime}. There are different aspects that may account for such an overhead. In our case, presumably, it is related to initialization time captured by the profiler and management cost for the embedded microprocessors reading out the results. However, as this overhead does not grow with the number of bits, it can be neglected when our adders are integrated into larger neuromorphic algorithms.



\section{Discussion}\label{sec.conclusion}

In this work we have described three designs for spike-based binary adders, and further implemented and tested them on Loihi 2 using the Lava framework. The implementations are readily available to be integrated as components into larger neuromorphic applications.

All three designs have their use cases. DCTA2 computes addition in the fastest manner, while using many synapses and having limited bit precision due to exponential synaptic weights. The sequential adder
supports high-precision addition using fewer \rev{spikes and synaptic operations, being more efficient in theory. This comes at the cost of runtime growing linearly with the number of bits. Further, in the specific case of Loihi 2, the theoretical advantage in terms of synapse count is outweighed by the reduced resources available due to the necessary use of high delays.} Compared to the other linear time algorithm~\cite{date_encoding_2023,wurm_arithmetic_2023}\rev{, which use the same delays in the implementation, our} sequential adder requires fewer resources, both in terms of neurons and synapses. DCTA3 provides a balance between DCTA2 and sequential adder regarding runtime, available precision, and synapse usage, at the cost of doubling the number of neurons.
\rev{Depending on the specific application and neuromorphic platform, the trade-off between runtime, hardware resources and energy consumption might be different. Energy consumption may be estimated based on chip characteristics and the numbers of spikes and synaptic operations reported in this work.}



While we focused here on adding non-negative integers, the right choice of encoding can extend the domain of these adders. As demonstrated by Wurm et al.~\cite{wurm_arithmetic_2023}, negative integers can be added if two's complement encoding is used. Furthermore, a different interpretation of bits can enable the addition of fixed-point numbers.

Other population codes for encoding integers in spiking neural networks offer alternatives to binary encoding. 
Binning assigns a neuron to each possible value in the representation, increasing the required space exponentially. Rate-based representations require many spikes to represent a single number, resulting in exponential time. With the same time requirements, single-spike-timing-based representation uses fewer spikes, thus reducing the energy required compared to the rate-based approach. While these representations are less efficient in time and use of neurons compared to binary representation, they offer other benefits such as increased robustness to errors in the input. In particular, a fault in the high bits can significantly change the interpretation of a binary number, which may be prevented by a rate-based representation. 
Moreover, when only a few operations are required and further processing before and after the addition requires another representation, it may be beneficial to use that representation for the addition and avoid the conversion.

The specific implementation provided is aimed at neuromorphic hardware systems with a fixed, not fully programmable, neuron model such as Loihi 1 (digital) or BrainScales-2 (analog), which can under certain circumstances enable extremely efficient emulation of spiking networks \cite{cramer2022surrogate, ostrau2022benchmarking}. These systems require neuron-based algorithms to increase the scope of problems they can solve. 
On fully programmable hardware like Loihi 2 or SpiNNaker 2, making direct use of the underlying processing architecture without adding the abstraction of neuron\rev{s} and synapses will be more efficient for standard arithmetic. However, as stated above, the minimal approach of hardware with fixed neuron models is promising, and therefore we aim to enhance its usability by implementing neuron-based arithmetic primitives. This vision is shared by others, constructing arithmetic logic units (ALUs) with multiple operations in a similar manner~\cite{ayuso-martinez_analog_2025}.



Through their parallel, efficient architecture, neuromorphic hardware systems offer a promising platform for algorithms in general, even beyond cognitive/biologically-constrained implementations~\cite{aimone_review_2022,lux2025hpc}.
Given the bottleneck in transferring data between neuromorphic chips and standard computers, solutions are required for on-chip computation to minimize data movement.
While data transfer rates and latency between neuromorphic systems will improve in the future, data transfer will remain a bottleneck compared to the increasing speed of computation. 
A mixture of neuro-inspired algorithms and on-chip implementations of standard tasks like data processing, analysis, and optimization, although perhaps computationally less efficient in time per operation, can minimize data movement and improve overall efficiency.

To achieve such implementations in spiking networks, multiple building blocks are required -- from basic logic gates~\cite{ayuso-martinez_analog_2025} to arithmetic operations such as addition and multiplication~\cite{aimone_spiking_2022,iaroshenko_binary_2023}. \rev{Implementations of such building blocks can take inspiration from neuroscience~\cite{mcculloch_logical_1943}, threshold logic~\cite{siu_depth-size_1991}, and traditional digital circuits.}
Adder circuits constitute a critical piece in this toolbox, which we have addressed by characterizing \rev{a set of} optimized, generally applicable neuromorphic adders.


\section*{Acknowledgments}
This work was funded by the Intel Corporation via a gift without restrictions and by the German Federal Ministry of Education and Research (BMBF) grant number 01IS22093A-E.
Elena C. Offenberg is now at the Department of Neurology and Neurosurgery, University Medical Center Utrecht.
Carlo Michaelis is now at the Department of Applied Information Technology, University of Gothenburg.
\section*{Author contributions}

Conceptualization: OS, JL, AL, CT; Data curation: OS; Formal analysis: OS, JL, AL; Funding acquisition: CM, JL, AL, CT; Investigation: OS, JL, AL; Methodology: OS, EO, CM, JL, AL; Project administration: JL, AL, CT; Software: OS; Resources: CT; Supervision: JL, AL, CT; Validation: OS; Visualization: OS, JL, AL; Writing -- original draft: OS, JL, AL; Writing -- review \& editing: all authors.

\bibliographystyle{IEEEtran}
\bibliography{intaddition.bib}

\begin{thebibliography}{10}
\providecommand{\url}[1]{#1}
\csname url@samestyle\endcsname
\providecommand{\newblock}{\relax}
\providecommand{\bibinfo}[2]{#2}
\providecommand{\BIBentrySTDinterwordspacing}{\spaceskip=0pt\relax}
\providecommand{\BIBentryALTinterwordstretchfactor}{4}
\providecommand{\BIBentryALTinterwordspacing}{\spaceskip=\fontdimen2\font plus
\BIBentryALTinterwordstretchfactor\fontdimen3\font minus
  \fontdimen4\font\relax}
\providecommand{\BIBforeignlanguage}[2]{{%
\expandafter\ifx\csname l@#1\endcsname\relax
\typeout{** WARNING: IEEEtran.bst: No hyphenation pattern has been}%
\typeout{** loaded for the language `#1'. Using the pattern for}%
\typeout{** the default language instead.}%
\else
\language=\csname l@#1\endcsname
\fi
#2}}
\providecommand{\BIBdecl}{\relax}
\BIBdecl

\bibitem{davies2021advancing}
M.~Davies, A.~Wild, G.~Orchard, Y.~Sandamirskaya, G.~A.~F. Guerra, P.~Joshi,
  P.~Plank, and S.~R. Risbud, ``Advancing neuromorphic computing with {Loihi}:
  A survey of results and outlook,'' \emph{Proceedings of the IEEE}, vol. 109,
  no.~5, pp. 911--934, 2021.

\bibitem{schuman2022opportunities}
C.~D. Schuman, S.~R. Kulkarni, M.~Parsa, J.~P. Mitchell, P.~Date, and B.~Kay,
  ``Opportunities for neuromorphic computing algorithms and applications,''
  \emph{Nature Computational Science}, vol.~2, no.~1, pp. 10--19, 2022.

\bibitem{sussillo_generating_2009}
D.~Sussillo and L.~F. Abbott, ``Generating {{Coherent Patterns}} of
  {{Activity}} from {{Chaotic Neural Networks}},'' \emph{Neuron}, vol.~63,
  no.~4, pp. 544--557, Aug. 2009.

\bibitem{ponulak2010supervised}
F.~Ponulak and A.~Kasi{\'n}ski, ``Supervised learning in spiking neural
  networks with {ReSuMe}: sequence learning, classification, and spike
  shifting,'' \emph{Neural Computation}, vol.~22, no.~2, pp. 467--510, 2010.

\bibitem{zenke2018superspike}
F.~Zenke and S.~Ganguli, ``{SuperSpike}: Supervised learning in multilayer
  spiking neural networks,'' \emph{Neural Computation}, vol.~30, no.~6, pp.
  1514--1541, 2018.

\bibitem{bellec2020solution}
G.~Bellec, F.~Scherr, A.~Subramoney, E.~Hajek, D.~Salaj, R.~Legenstein, and
  W.~Maass, ``A solution to the learning dilemma for recurrent networks of
  spiking neurons,'' \emph{Nature communications}, vol.~11, no.~1, p. 3625,
  2020.

\bibitem{aimone_provable_2021}
J.~B. Aimone, Y.~Ho, O.~Parekh, C.~A. Phillips, A.~Pinar, W.~Severa, and
  Y.~Wang, ``Provable {{Advantages}} for {{Graph Algorithms}} in {{Spiking
  Neural Networks}},'' in \emph{Proceedings of the 33rd {{ACM Symposium}} on
  {{Parallelism}} in {{Algorithms}} and {{Architectures}}}, ser. {{SPAA}}
  '21.\hskip 1em plus 0.5em minus 0.4em\relax New York, NY, USA: Association
  for Computing Machinery, 2021, pp. 35--47.

\bibitem{HennessyPatterson2012}
J.~L. Hennessy and D.~A. Patterson, \emph{Computer architecture: a quantitative
  approach (with online appendix)}, 5th~ed.\hskip 1em plus 0.5em minus
  0.4em\relax Elsevier, Waltham/MA, USA, 2012.

\bibitem{ramos_two_1999}
J.~Ramos and A.~Boh\'{o}rquez, ``Two operand binary adders with threshold
  logic,'' \emph{IEEE Transactions on Computers}, vol.~48, no.~12, pp.
  1324--1337, 1999.

\bibitem{siu_depth-size_1991}
K.~Siu, V.~Roychowdhury, and T.~Kailath, ``Depth-size tradeoffs for neural
  computation,'' \emph{IEEE Transactions on Computers}, vol.~40, no.~12, pp.
  1402--1412, 1991.

\bibitem{date_encoding_2023}
P.~Date, S.~Kulkarni, A.~Young, C.~Schuman, T.~Potok, and J.~Vetter, ``Encoding
  integers and rationals on neuromorphic computers using virtual neuron,''
  \emph{Scientific Reports}, vol.~13, no.~1, p. 10975, 2023.

\bibitem{wurm_arithmetic_2023}
A.~Wurm, R.~Seay, P.~Date, S.~Kulkarni, A.~Young, and J.~Vetter, ``Arithmetic
  {{Primitives}} for {{Efficient Neuromorphic Computing}},'' in \emph{2023
  {{IEEE International Conference}} on {{Rebooting Computing}} ({{ICRC}})},
  2023, pp. 1--5.

\bibitem{aimone_spiking_2022}
J.~B. Aimone, A.~J. Hill, W.~M. Severa, and C.~M. Vineyard, ``Spiking neural
  streaming binary arithmetic,'' in \emph{2021 International Conference on
  Rebooting Computing (ICRC)}, 2021, pp. 79--83.

\bibitem{iaroshenko2023binary}
O.~Iaroshenko, A.~T. Sornborger, and D.~C. Arana, ``Binary operations on
  neuromorphic hardware with application to linear algebraic operations and
  stochastic equations,'' \emph{Neuromorphic Computing and Engineering},
  vol.~3, no.~1, p. 014002, 2023.

\bibitem{ayuso-martinez_analog_2025}
A.~{Ayuso-Martinez}, D.~{Casanueva-Morato}, J.~P. {Dominguez-Morales},
  G.~Indiveri, A.~{Jimenez-Fernandez}, and G.~{Jimenez-Moreno}, ``Analog
  {{Implementation}} of a {{Spiking System}} for {{Performing Arithmetic Logic
  Operations}} on {{Mixed-Signal Neuromorphic Processors}},'' \emph{Advanced
  Intelligent Systems}, vol.~7, no.~4, p. 2400524, 2025.

\bibitem{timcheck2023intel}
J.~Timcheck, S.~B. Shrestha, D.~B.~D. Rubin, A.~Kupryjanow, G.~Orchard,
  L.~Pindor, T.~Shea, and M.~Davies, ``The {I}ntel neuromorphic {DNS}
  challenge,'' \emph{Neuromorphic Computing and Engineering}, vol.~3, no.~3, p.
  034005, 2023.

\bibitem{hala-point-news}
\BIBentryALTinterwordspacing
``Intel builds world's largest neuromorphic system to enable more sustainable
  {AI},'' 2024, accessed: 14 November 2024. [Online]. Available:
  \url{https://www.techpowerup.com/321645/intel-builds-worlds-largest-neuromorphic-system-to-enable-more-sustainable-ai}
\BIBentrySTDinterwordspacing

\bibitem{lava-website}
\BIBentryALTinterwordspacing
``Lava software framework,'' 2024, accessed: 11 November 2024. [Online].
  Available: \url{https://lava-nc.org/}
\BIBentrySTDinterwordspacing

\bibitem{orchard2022talk}
G.~Orchard, ``{Understanding the Loihi 2 architecture},'' 2022, presentation at
  Intel Neuromorphic Research Community (INRC) Fall Workshop 2022.

\bibitem{cramer2022surrogate}
B.~Cramer, S.~Billaudelle, S.~Kanya, A.~Leibfried, A.~Gr{\"u}bl, V.~Karasenko,
  C.~Pehle, K.~Schreiber, Y.~Stradmann, J.~Weis \emph{et~al.}, ``Surrogate
  gradients for analog neuromorphic computing,'' \emph{Proceedings of the
  National Academy of Sciences of the USA}, vol. 119, no.~4, p. e2109194119,
  2022.

\bibitem{ostrau2022benchmarking}
C.~Ostrau, C.~Klarhorst, M.~Thies, and U.~R{\"u}ckert, ``Benchmarking
  neuromorphic hardware and its energy expenditure,'' \emph{Frontiers in
  Neuroscience}, vol.~16, p. 873935, 2022.

\bibitem{aimone_review_2022}
J.~B. Aimone, P.~Date, G.~A. {Fonseca-Guerra}, K.~E. Hamilton, K.~Henke,
  B.~Kay, G.~T. Kenyon, S.~R. Kulkarni, S.~M. Mniszewski, M.~Parsa, S.~R.
  Risbud, C.~D. Schuman, W.~Severa, and J.~D. Smith, ``A review of
  non-cognitive applications for neuromorphic computing,'' \emph{Neuromorphic
  Computing and Engineering}, vol.~2, no.~3, p. 032003, 2022.

\bibitem{lux2025hpc}
N.~Lux, C.~Mandal, J.~Decker, J.~Luboeinski, J.~Drewljau, T.~Meisel,
  C.~Tetzlaff, C.~Boehme, and J.~Kunkel, ``{HPC-AI} benchmarks – a
  comparative overview of high-performance computing hardware and {AI}
  benchmarks across domains,'' \emph{Journal of Artificial Intelligence and
  Robotics}, vol.~1, no.~2, 2024.

\bibitem{iaroshenko_binary_2023}
O.~Iaroshenko, A.~T. Sornborger, and D.~C. Arana, ``Binary operations on
  neuromorphic hardware with application to linear algebraic operations and
  stochastic equations,'' \emph{Neuromorphic Computing and Engineering},
  vol.~3, no.~1, p. 014002, Jan. 2023.

\bibitem{mcculloch_logical_1943}
W.~S. McCulloch and W.~Pitts, ``A logical calculus of the ideas immanent in
  nervous activity,'' \emph{The bulletin of mathematical biophysics}, vol.~5,
  no.~4, pp. 115--133, Dec. 1943.

\end{thebibliography}
\vspace{12pt}

\end{document}